\documentclass[review]{elsarticle}

\usepackage{multirow}
\usepackage{hyperref}
\usepackage{lineno}

\begin{document}

\begin{frontmatter}

\title{Predicting the time-evolution of multi-physics systems with sequence-to-sequence models}

\author[llnl,tamu]{K. D. Humbird}
\ead{humbird1@llnl.gov}

\author[llnl]{J. L. Peterson}
\author[nd]{R. G. McClarren}

\address[llnl]{Lawrence Livermore National Laboratory, 7000 East Ave, Livermore, CA 94550}
\address[tamu]{Department of Nuclear Engineering, Texas A \& M University, 3133 TAMU, College Station, TX 77843}
\address[nd]{Department of Aerospace and Mechanical Engineering, University of Notre Dame, 365 Fitzpatrick Hall, Notre Dame, IN, 46556}

\begin{abstract}
In this work, sequence-to-sequence (seq2seq) models, originally developed for language translation, are used to predict the temporal evolution of complex, multi-physics computer simulations. The predictive performance of seq2seq models is compared to state transition models for datasets generated with multi-physics codes with varying levels of complexity-- from simple 1D diffusion calculations to simulations of inertial confinement fusion implosions. Seq2seq models demonstrate the ability to accurately emulate complex systems, enabling the rapid estimation of the evolution of quantities of interest in computationally expensive simulations.
\end{abstract}


\end{frontmatter}


\section{Introduction}
Computer simulations of detailed multi-physics systems often take several hours to run, making exploration throughout a vast design space prohibitively expensive. A common method for mapping design spaces of large computer codes is to train a machine learning model to emulate the code in a region of the design space \citep{surrogate1,surrogate2,surrogate3}. The machine learning model, often called a ``surrogate'', learns to accurately interpolate between a set of simulations spread throughout the reduced design space, such that it can rapidly predict quantities of interest anywhere within that space without requiring additional expensive simulations. 

Feed forward neural networks (FFNNs) have shown particularly good performance as surrogate models for expensive multi-physics codes \cite{nnphysics, djinn, nnphysics2}. However, FFNNs are best-suited for predicting quantities of interest that are fixed-length vectors or arrays; they are not formulated for time series or sequence data of arbitrary length. In many multi-physics simulations, time series data or data from discretized meshes often varies in size between simulations depending on the boundary conditions. For example, in hydrodynamics codes, simulations with slightly different initial conditions can require a different number of time steps and spatial discretization cells, as the resolution of the simulation can dynamically change to ensure physical processes are modeled correctly \cite{pomraning_rh,mihalas_radh}.

FFNNs have been successfully applied to time series data by training the model to learn the ``state transition'' for the quantities of interest (QOIs) from one time step to the next \cite{statetrans}. The model can be iterated upon to predict the entire evolution of QOIs by using the prediction from one time step as the input to the model for the next time step.  A trained state transition model can thus be given the initial conditions of a system and predict the entire trajectory of the QOIs. Although this allows for predictions of arbitrary sequence length, state transition models are incapable of learning long-term relationships in the data.

Recurrent neural networks (RNNs) are simple generalizations of FFNNs that allow the network to retain memory of its previous states, making them attractive options for modeling sequential data \cite{rnn}. In an RNN, neurons are replaced with ``cells'', which take in an input vector at the current sequence step, like a traditional neuron, but also take in the current cell state such that the cell can learn dependencies between data at different steps in the sequence. 
Gated recurrent units (GRU)~\cite{gru,gru2}, a variant of the long short-term memory (LSTM) cell~\cite{lstm,lstmslidingtime} are commonly used cell structures as they are particularly well-suited for learning long-range dependencies in the data. The GRU is a variant of the standard recurrent network that has an update gate, which determines how much information from the past needs to be passed along toward the future, and a reset gate, which determines how much previous information to forget. 
RNNs are capable of mapping arbitrary length input sequences to arbitrary length output sequences. These models are applied to a wide range of problems, from image captioning to machine translation \cite{conversation,imgcapt,imgcapt2}. Recently, RNN models have been used to predict the time evolution of fluid dynamics simulations~\cite{fluids} by combining convolutional layers with LSTM cells. In this work, we focus on sequence-to-sequence (seq2seq) architectures, originally applied to machine translation~\cite{s2s,s2smulti}, to create surrogate models of multi-physics systems. 

Standard RNN architectures can map input sequences to output sequences when the alignment between inputs and outputs is known; however, it is unclear how to handle situations in which the input and output sequences have lengths which differ from one example to the next within the same set of training data, such as variable time step simulation data. Seq2seq models handle variable sequence lengths by mapping an input sequence into a fixed-length ``latent'' vector via an ``encoder'' network. The latent vector is then mapped into a variable length output sequence via a ``decoder''. For example, in language translation the input sequence could be an English phrase, which is encoded into a latent ``thought vector'' that captures the meaning of the phrase, and is then decoded into the equivalent French phrase. For time series data, the input sequence is the first several time steps in the evolution of a system, which gets encoded into the fixed-dimensional latent space.  The latent space representation of the data is decoded to produce a series of predictions for the subsequent evolution of the system. 

\begin{figure}[h]
\begin{center}
		\includegraphics[width=0.85\textwidth]{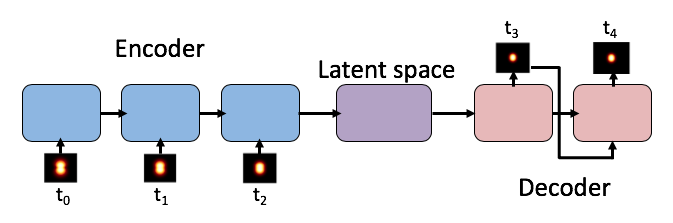}
		\caption{Unrolled seq2seq architecture. The blue and red cells represent stacks of recurrent cells (such as GRUs) that make up the encoder and decoder, respectively. The input image time series is compressed into a latent vector via the encoder, which is then decoded to make a prediction for the output image time series.}
		\label{fig:s2sp}
\end{center}
\end{figure}

The architecture of a seq2seq model, unrolled to illustrate each step of the input and output sequences, is shown in Fig. \ref{fig:s2sp}, where the blue and red cells represent a stack of LSTM or GRU cells for the encoder and decoder, respectively~\cite{s2s}. The observed portion of the time series is input to the encoder (blue cells), and the cell state is passed from one time step to the next until this information arrives at the latent space -- the fixed-dimensional vector that represents the entire input time series. The latent vector is then decoded to produce the output prediction for the unobserved portion of the time series (red cells). The decoder predicts the output sequence one unit at a time, using its prediction for the current values of the QOIs and the cell state as inputs to predict the next value of the QOIs.

The ability of seq2seq models to handle arbitrary and variable length sequential data makes them attractive for emulating dynamic multi-physics simulations that evolve in a variable number of time steps within the same data set. Such simulations can be computationally expensive; emulators that can be evaluated rapidly are attractive for applications such as design optimization or hypothesis testing. 

We are particularly interested in using seq2seq models to efficiently explore the parameter space of computationally expensive multi-physics codes. In this application, a seq2seq model trained on a set of simulations that are scattered throughout the parameter space is evaluated on-the-fly while running simulations at new locations. If the seq2seq model predicts the time evolution of the QOIs in the new simulations accurately, the predictable simulations are terminated in flight, and computational resources are allocated to regions of parameter space that are less predictable; this can result in significant computational savings for expensive simulations. This idea has been explored with state transition models \cite{victalk}, in which the author efficiently trained a state transition model by terminating simulations that were predictable, and investing computer resources in areas of parameter space where the state transition model did not perform well. This intelligent sampling of design space is a simple extension to adaptive learning techniques \cite{adapatlearn,adaptlearn2,adaptlearn3} which, in addition to resolving the response surface for end-of-time QOIs, also learns the time evolution of the response surface. 

In this work, we evaluate the ability of seq2seq models to accurately emulate dynamic multi-physics simulations. We compare seq2seq models to state transition models for datasets generated with multi-physics codes that vary in complexity, and produce QOIs with dimensionality that spans several orders of magnitude. The datasets are introduced in section 2.1, followed by the performance comparison in section 2.2. The disadvantages of state transition models are highlighted, along with the benefits seq2seq models offer for high-dimensional QOIs. Applications of seq2seq models for physics simulation data are discussed in section 3, including the benefits seq2seq models could offer for advanced sampling techniques, and how they can be applied to numerical convergence studies. 

\section{Predicting the time evolution of multi-physics systems}
In the following sections, seq2seq models are compared to neural network state transition models for three datasets generated with computer simulations of varying complexity. The multi-physics codes and corresponding datasets are presented in section 2.1, followed by the performance comparison in section 2.2.

\subsection{Databases}
Seq2seq models are compared to state transition models for three sets of simulation data of increasing complexity. The databases are generated using a 1D time-dependent diffusion code, a 1D Lagrangian radiation-hydrodynamics code, and a 3D multi-physics code that simulates inertial confinement fusion implosions. Along with the increasing complexity of the underlying physics models, the outputs generated in each code also become more complicated. In the following subsections, each dataset and the codes used to generate them are described in detail.


\subsubsection{1D diffusion}
The diffusion database contains solutions to a simple 1D diffusion problem that determines the spatially-integrated concentration in a system as a function of time. The time evolution of the concentration is found by solving a discretized diffusion equation on a grid of arbitrary resolution. The resolution of the grid has a significant impact on the accuracy of the solution; a coarse grid will have high error, and as the grid is refined the solution will approach the true answer at an asymptotic rate. 

To generate a database of diffusion solutions, the problem is set up as follows: the spatially-integrated concentration as a function of time is solved for in a 1D slab of unit length with a constant diffusion coefficient $D$. There is a source with value unity at the left boundary, which is turned on at time $t$=0. This problem has an analytic solution, given by Eq. \ref{eq1}: 

\begin{equation}\label{eq1}
y(t)=1-2\sqrt{Dt/\pi} \cdot e^{1/(2\sqrt{Dt})^2}-\mathrm{erf}\left( \frac{1}{2\sqrt{Dt}} \right).
\end{equation}

An implicit finite difference solver (centered-difference in space, backward Euler in time ~\cite{numericalrecipes}) is used to compute the total concentration for varying values of the diffusion coefficient ($D$, between 1.0 and 3.0) and the spatial discretization step size ($dx$ between 1.0e-3 and 1.0e-5). The spatially-integrated concentration as a function of time is recorded for 1000 time steps (step size of 1.0e-6) for 1000 random combinations of $D$ and $dx$.

\subsubsection{1D radiation hydrodynamics}
The radiation hydrodynamics dataset tests the ability of the models to predict high-dimensional time histories generated by a more expensive multi-physics code with non-analytic solutions. In this task, the models are trained to predict the space and time evolution of multiple QOIs for systems described by the radiation hydrodynamics equations with grey diffusion (RHGD) \cite{pomraning_rh}. 

A Lagrangian RHGD code (\textit{unpublished results}) is used to generate solutions on a 1D grid of 100 spatial cells for 100 time steps. The problem domain is a 1D slab with an initial low temperature throughout the domain and applied to the boundaries. At time $t$=0 a higher temperature source on the left boundary is turned on. To create the database of simulations, three parameters are varied to generate 500 Latin hypercube-sampled \cite{lhs} points: the initial domain temperature is varied between 1.0e-6 and 0.01, the initial density of the slab between 0.75 and 1.25, and the temperature of the source on the left boundary of the slab between 0.1 and 0.6. The seq2seq models are trained to predict the evolution of the temperature, pressure, and density as a function of time and position on the grid, totaling 300 time-varying parameters. 

\subsubsection{3D inertial confinement fusion implosion simulations}
The final dataset used to compare state transition and seq2seq models is generated with a semi-analytic inertial confinement fusion (ICF) implosion code \cite{jim1,jim2}. The implosion model describes the stagnation of an ICF implosion using an analytic spatial transport model, a 3-D shape model, and a numerical energy balance model. The spatial approximations allow an efficient description of the implosion without need for a numerical spatial mesh, while retaining the zeroth-order spatial information. This allows the efficient prediction of a large set of experimental observables including images, time series and scalars. 

Three parameters are varied in the ICF model to create a database of 300 simulations. The parameters are applied shape asymmetry multipliers expressed as spherical harmonic modes ($Y_l^m$ for $l$=2, $m$=[-1,0,1]) which vary in magnitude from [1.0,1.5], and degrade the implosion performance by distorting the shell of the ICF fuel capsule. 
The code generates multiple outputs; for this application the primary QOI is the X-ray emission image of the imploding capsule as a function of time. The images are 64x64 pixels with one color channel, and are recorded at 10 points in time throughout the implosion. 

\subsection{Comparison of seq2seq and state transition models}
The three datasets described above are used to train state transition models and seq2seq models to predict the time evolution of the QOIs given only the initial conditions. Seq2seq models receive an ``input'' sequence that contains the initial condition only (the simulation inputs and the QOIs at the first time step), and the ``output'' sequence is the subsequent evolution of the QOIs (the simulation inputs, which are unchanged, and the evolving QOIs for the remaining time steps). The state transition models are trained using input vectors that include the simulation inputs and the QOIs at a time step $t$, and the corresponding output vectors contain the QOIs at time step $t+1$. For evaluation, the state transition model is provided the simulation inputs and the initial value of the QOIs to predict the state at the first time step, and this solution is iterated upon to generate the complete time series.

The state transition model is a fully connected ``deep jointly-informed neural network'' (DJINN) \cite{djinn}. The DJINN models are trained with optimized hyper-parameters chosen by the software; these settings are summarized in Table \ref{table:results}. 
The seq2seq model architectures are similar for all three datasets. The encoder and decoder share the same architecture, which is composed of two stacked GRU cells with a fixed number of hidden units per cell. The learning rate, batch size, and number of training epochs are chosen to optimize the performance on the test dataset; these parameters are also summarized in Table \ref{table:results}. 

Each of the three datasets is split into an 80\% training set and 20\% test set, the latter is used to compare the performance of the two models. For the state transition models, the full sequences are sectioned into two time step slices (transitions), with time step $t$ as an input, and $t+1$ as the corresponding output. For the seq2seq models, each sequence is a single training example with time step $t=0$ as the input, and time steps $t=[1, t_{final}]$ as the output sequence. Each dataset is scaled between [0,1] prior to training the networks. 

To compare the performance of the two models, the integrated absolute error is computed, normalized by the sequence length, $N_s$:

\begin{equation}\label{eq:IAE}
IAE = \frac{1}{N_s} \sum_{n=1}^{N_s} |Y_{n}^{true}-Y_{n}^{pred}|,
\end{equation}

where $Y_n^{pred}$ and $Y_n^{true}$ are the predicted and true values of the QOI at sequence step $n$, respectively. Reported in Table \ref{table:results} are the mean, median, and standard deviation for test set IAE. 

\begin{table*}[]
\caption{Model hyper-parameters and performance metrics. The state transition model, DJINN, is a fully connected neural network with the neurons per hidden layer specified in the architecture column (this excludes input and output layers). The seq2seq models are composed of two stacked GRU cells with the number of hidden units per cell specified in the architecture column. The batch size is specified in units of transitions (trans) for the state transition model, and sequences (seq) for the seq2seq model.}

\vspace{5mm}
\label{table:results}
\resizebox{\textwidth}{!}{
\begin{tabular}{c|c|c|c|c|c|c|c|c}
Dataset    & Model   & Architecture    & Learn rate & Batch size & \# Epochs & Mean IAE & Median IAE & SD IAE \\ \hline
\multirow{2}{*}{Diffusion} & DJINN   & 4, 8, 13     & 0.006     & 4000 trans & 400    & 0.0329   & 0.0327     & 0.0059 \\
                           & seq2seq & 14 units    & 0.001      & 20 seq    & 3000    & 0.0257   & 0.0243     & 0.0086 \\ \hline
\multirow{2}{*}{Rad-hydro} & DJINN   & 302, 159   & 0.002    & 2000 trans & 400    & 0.0632   & 0.0607     & 0.0155 \\
                           & seq2seq & 26 units   & 0.001   & 10 seq    & 5500   & 0.0419   & 0.0391     & 0.0086 \\ \hline
\multirow{2}{*}{ICF}       & DJINN   & 1970, 2006    & 0.0004     &  72 trans   & 100   & 0.0098   &   0.0085   & 0.0035    \\
                           & seq2seq & 16 units   & 0.001  & 30 seq    & 1500      & 0.0046   & 0.0037     & 0.0020
\end{tabular}}
\end{table*}
\vspace{5mm}

\begin{figure}[h]
\begin{center}
		\includegraphics[width=0.75\textwidth]{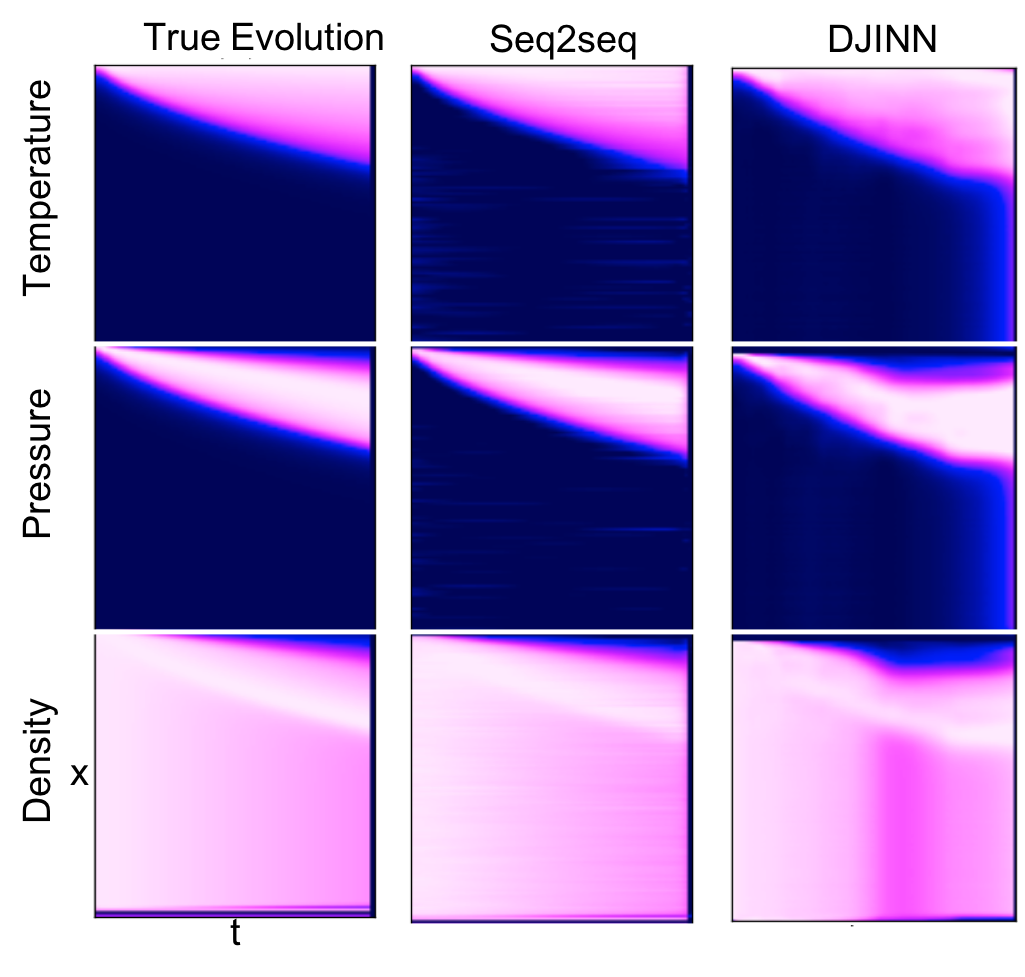}
		\caption{Example prediction from the radiation hydrodynamics dataset. Color maps on all panels are scaled between [0,1]. The true evolution of the system is illustrated on the left, followed by the predictions from the seq2seq (middle) and state transition DJINN model (right). The seq2seq model produces a smoother and more accurate evolution than the state transition model.}
		\label{fig:rh}
\end{center}
\end{figure}

The seq2seq model consistently out-performs the state transition model, particularly as the dimensionality of the QOIs increases. To illustrate the quality of predictions from the seq2seq and state transition models, Figure \ref{fig:rh} shows an example from the radiation hydrodynamics dataset. The seq2seq model predicts a smoother and more accurate evolution of the system than the DJINN state transition model, which displays visible discontinuities between time steps. The recurrent connections in the seq2seq decoder allow the model to process the entire output sequence simultaneously and leverage long and short-term correlations in the data, resulting in a smooth prediction for the temporal and spatial evolution of the system. 

\begin{figure*}
\begin{center}
		\includegraphics[width=1.\textwidth]{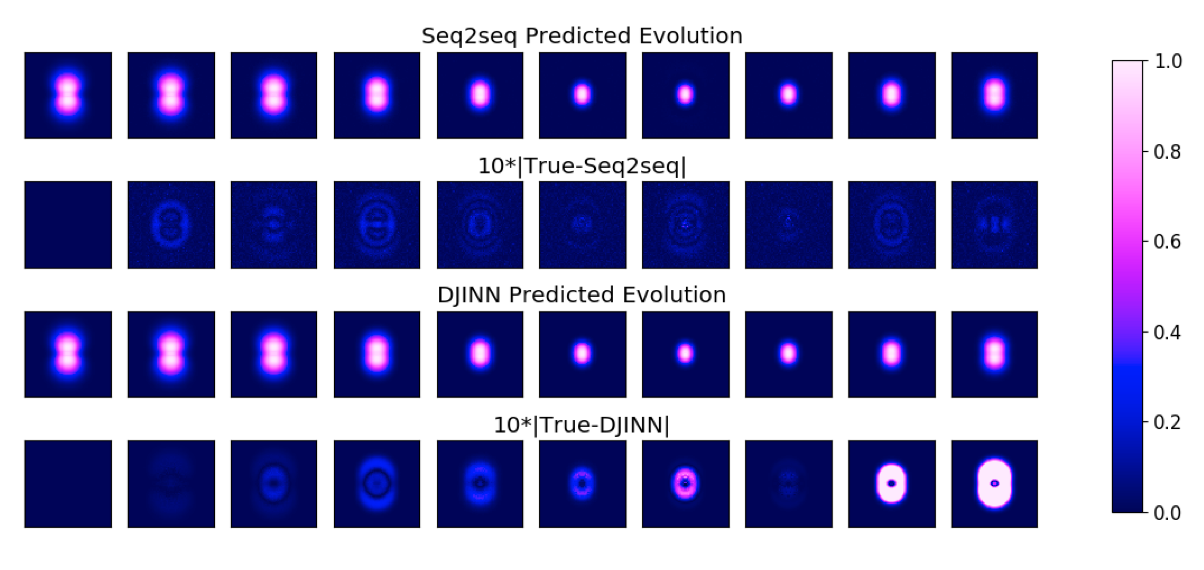}
		\caption{Example predictions from the ICF database. The first image in the first and third row is the model input, and the subsequent nine images illustrate the evolution as predicted by the seq2seq and DJINN models, respectively. The second and fourth rows show the absolute error in each model's predictions multiplied by a factor of 10 for visibility. The seq2seq model has lower error than the DJINN state transition model; both models display highest error near steep gradients.}
		\label{fig:jag}
\end{center}
\end{figure*}

The ICF X-ray emission image dataset includes the highest number of input dimensions (4096 pixels) and the training data is limited to only 240 complete time series. The DJINN state transition model performs significantly worse than the seq2seq model, as fully-connected networks are not well-suited for high-dimensional image data unless preceded by convolutional layers or other dimensional reduction techniques \cite{autoencoder, cnn, cnn2}. 
Fig. \ref{fig:jag} shows an example seq2seq and DJINN prediction for this dataset. The seq2seq model demonstrates low integrated error, but displays regions of up to 20\% prediction error near steep gradients in the pixel values; additional training data may be required to reduce the error further. The DJINN state transition model also suffers near high gradients, and displays particularly high error in the final frames of the sequence. The first seven transitions in each sequence show the X-ray emission region compressing, and only the final two frames illustrate the subsequent expansion of the emission region. The state transition model, which only learns the transition between frames, is therefore much better at predicting the first several frames of the sequence, as the model has been exposed to significantly more data that illustrates compression from one state to the next. The seq2seq model, however, is able to predict the compression and expansion equally well, as this model learns the full evolution of the sequence, and can thus recognize that the X-ray emission region often expands toward the end of the sequence.

\section{Using seq2seq models to improve physics simulations}
\begin{figure}[h]
\begin{center}
		\includegraphics[width=0.75\textwidth]{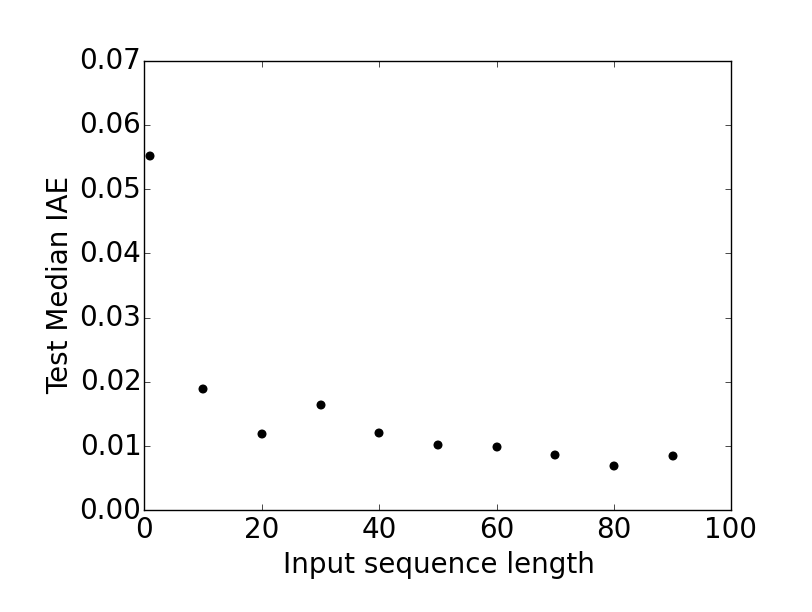}
		\caption{Median IAE for the test dataset as a function of the input sequence length. The model predicts the evolution with lower error per time step as the length of the input sequence increases. At an input sequence length of 50, the rate of the error reduction slows, suggesting the model only needs to observe half of the evolution before it can predict the remainder of the trajectory with minimal error.}
		\label{fig:err}
\end{center}
\end{figure}

Seq2seq models are attractive for multi-physics codes in which simulations can produce data of variable sequence length within a single database, as they readily handle variable length input and output sequences. As mentioned in section 1, this enables a trained seq2seq model to predict the evolution of a system at various points in time as a simulation is progressing. For example, a simulation runs for 10 time steps, passes the data to a seq2seq model to predict the next 10 time steps, and once the simulation reaches time step 20, the error in the model's prediction is evaluated. If the seq2seq model is accurately predicting the evolution of the system, computational time can be saved by stopping the simulation early, and relying on the seq2seq model to predict the remainder of the evolution. It is reasonable to expect that there is a minimum number of time steps the simulation must complete, at which point the seq2seq model can accurately predict the remainder of the evolution. This idea is tested on the diffusion dataset. The 800 time histories from the training dataset are split into random-length input and output sequences, both varying in length from 10 to 90 time steps. The model is trained on the variable sequence length data, and is evaluated on the test dataset, which has a fixed input sequence length. To determine the optimal input sequence length required to accurately predict the remainder of the diffusion solution, the fixed input sequence length for the test data is varied from 10 to 90 time steps. The test error as a function of input sequence length is shown in Fig. \ref{fig:err}.  

As expected, the average prediction error per time step decreases as the sequence length of the prediction decreases. The rate of error reduction slows around an input sequence length of 50 time steps, suggesting that it is sufficient to run half of a new diffusion simulation, at which point the seq2seq model can predict the remainder of the evolution with minimal error. In the case of the diffusion code which can be executed quickly, the computational time savings gained by stopping the simulation halfway through is small. However, for expensive multi-physics codes that take several hours to run, the amount of computational time saved by training a seq2seq model can be significant. Future work will explore the possibility of saving computational time required to simulate complex dynamic systems by training a seq2seq emulator to complete trajectories of QOIs given only a small fraction of the system's evolution.

\begin{figure}[h]
\begin{center}
		\includegraphics[width=0.9\textwidth]{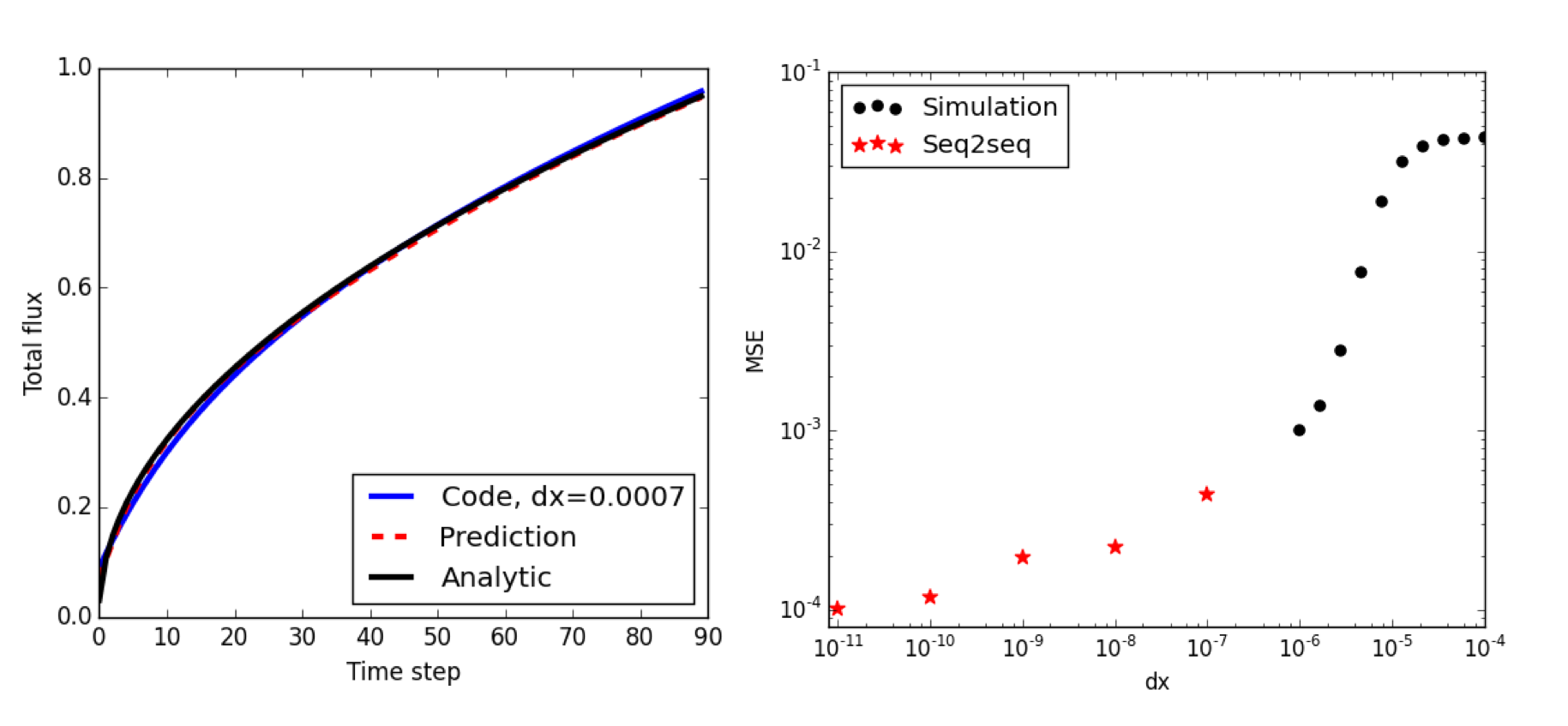}
		\caption{Left: normalized total concentration for the 1D diffusion problem with a diffusion coefficient of 1.34. The black line shows the analytic solution, the blue is the numerical solution with $dx$=7$\times10^{-4}$, and the red line is the prediction of the seq2seq model with $dx$=0. Right: the error of the finite element solution as $dx$ is decreased is shown in black; the red stars indicates the seq2seq prediction error (compared to the analytic solution) as dx is decreased toward 0.}
		\label{fig:diff}
\end{center}
\end{figure}

Another interesting application of the diffusion seq2seq model is its ability to predict solutions for various values of the discretization resolution, $dx$, given only the initial conditions of the system. Although it requires extrapolation, the model is evaluated with increasingly small $dx$ and compared to the analytic solution. In Figure \ref{fig:diff}, the predicted, analytic, and numerical solutions with a finite $dx$ are shown on the left for $D$=1.34. On the right, the black points indicate the error in the numerical solution as a function of $dx$ is computed for the same value of $D$, which shows the expected convergence rate of $dx^2$ for small $dx$. The stars indicates the seq2seq model error as $dx$ is decreased beyond the boundary of the training data. The seq2seq model predictions for small $dx$ do not follow the second order convergence rate of the numerical solution, but the predictions have significantly lower error than solutions from the training data, suggesting the model does have a limited ability to extrapolate. 

In this simple example, the diffusion equation can be solved quickly on a high resolution grid to estimate the converged solution without relying on seq2seq models. However, for more complicated systems it might be necessary to run a set of simulations with increasing levels of resolution to estimate the convergence rate of the solution. The data generated in the convergence study can be used to train a seq2seq model which can estimate converged solutions for systems which cannot be solved analytically. 

\section{Conclusions}
In this work, we have demonstrated the ability of seq2seq models to predict the temporal evolution multi-physics simulations of varying complexity. The encoder-decoder structure with recurrent connections offers many advantages over simple state transition models: the seq2seq models can learn long-term dependencies in data, can handle arbitrary sequence lengths, and can process high dimensional quantities of interest effectively. By learning to accurately emulate multi-physics design codes, seq2seq models enable rapid estimation of the time trajectory of quantities of interest at a fraction of the cost of a full simulation. 

\section*{Acknowledgements}
The authors would like to thank Jim Gaffney for access to the ICF implosion code, and Vic Castillo and Brian Spears for discussions on intelligent sampling. 
This work was performed under the auspices of the U.S. Department of Energy by Lawrence Livermore National Laboratory under Contract DE-AC52-07NA27344. Released as LLNL-JRNL-756184-DRAFT.
This document was prepared as an account of work sponsored by an agency of the United States government. Neither the United States government nor Lawrence Livermore National Security, LLC, nor any of their employees makes any warranty, expressed or implied, or assumes any legal liability or responsibility for the accuracy, completeness, or usefulness of any information, apparatus, product, or process disclosed, or represents that its use would not infringe privately owned rights. Reference herein to any specific commercial product, process, or service by trade name, trademark, manufacturer, or otherwise does not necessarily constitute or imply its endorsement, recommendation, or favoring by the United States government or Lawrence Livermore National Security, LLC. The views and opinions of authors expressed herein do not necessarily state or reflect those of the United States government or Lawrence Livermore National Security, LLC, and shall not be used for advertising or product endorsement purposes.

\section*{References}

\bibliography{bib}

\end{document}